\documentclass[runningheads]{llncs}
\usepackage{graphicx}

\usepackage{tikz}
\usepackage{comment}
\usepackage{amsmath,amssymb} 
\usepackage{color}

\usepackage{epsfig}
\usepackage{graphicx}
\usepackage{amsmath}
\usepackage{amssymb}         
\usepackage{floatrow}
\usepackage{wrapfig}

\usepackage{soul}
\usepackage{dsfont}
\usepackage{multirow}

\usepackage{bbm}
\usepackage{diagbox}
\usepackage{makecell}
\usepackage{adjustbox}

\usepackage{algorithm}
\usepackage{algpseudocode}

\usepackage[accsupp]{axessibility}  

\begin{document}
\pagestyle{headings}  
\mainmatter
\def\ECCVSubNumber{3060}  

\title{Exploring Hierarchical Graph Representation for Large-Scale Zero-Shot Image Classification}


\titlerunning{HGR-Net}
%
\author{Kai Yi\inst{1} \and
Xiaoqian Shen\inst{1} \and
Yunhao Gou\inst{12} \and
Mohamed Elhoseiny\inst{1}}
\authorrunning{Kai Yi et al.}
%
\institute{King Abdullah University of Science and Technology (KAUST) \and
University of Electronic Science and Technology of China (UESTC)\\
\email{\{kai.yi, xiaoqian.shen, yunhao.gou, mohamed.elhoseiny\}@kaust.edu.sa}}
\maketitle
  
\begin{abstract}
The main question we address in this paper is how to scale up visual recognition of unseen classes, also known as zero-shot learning, to tens of thousands of categories as in the ImageNet-21K benchmark. At this scale, especially with many fine-grained categories included in ImageNet-21K, it is critical to learn quality visual semantic representations that are discriminative enough to recognize unseen classes and distinguish them from seen ones. We propose a \emph{H}ierarchical \emph{G}raphical knowledge \emph{R}epresentation framework for the confidence-based classification method, dubbed as HGR-Net. Our experimental results demonstrate that HGR-Net can grasp class inheritance relations by utilizing hierarchical conceptual knowledge. Our method significantly outperformed all existing techniques, boosting the performance by 7\% compared to the runner-up approach on the ImageNet-21K benchmark. We show that HGR-Net is learning-efficient in few-shot scenarios. We also analyzed our method on smaller datasets like ImageNet-21K-P, 2-hops and 3-hops, demonstrating its generalization ability. Our benchmark and code are available at {https://kaiyi.me/p/hgrnet.html}.  
\keywords{zero-shot learning, semantic hierarchical graph, large-scale knowledge transfer, vision and language}
\end{abstract}

\section{Introduction}
Zero-Shot Learning (ZSL) is the task of recognizing images from unseen categories with the model trained only on seen classes. Nowadays, ZSL relies on semantic information to classify images of unseen categories and can be formulated as a visual semantic understanding problem. In other words, given candidate text descriptions of a class that has not been seen during training, the goal is to identify images of that unseen class and distinguish them from seen ones and other unseen classes based on their text descriptions.

In general, current datasets contain two commonly used semantic information including attribute descriptions (e.g., AWA2~\cite{gbu}, SUN~\cite{patterson2012sun}, and CUB~\cite{welinder2010caltech}), and more challenging unstructured text descriptions (e.g., CUB-wiki\cite{elhoseiny2013write}, NAB-wiki~\cite{elhoseiny2017link}). However, these datasets are all small or medium-size with up to a few hundred classes, leaving a significant gap to study generalization at a realistic scale. In this paper, we focus on large-scale zero-shot image classification. More specifically, we explore the learning limits of a model trained from 1K seen classes and transfer it to recognize more than 10 million images from 21K unseen candidate categories from ImageNet-21K~\cite{deng2009imagenet}, which is the largest available image classification dataset to the best of our knowledge. 

\begin{figure}[!t]
    \centering
    \includegraphics[width=0.7\textwidth]{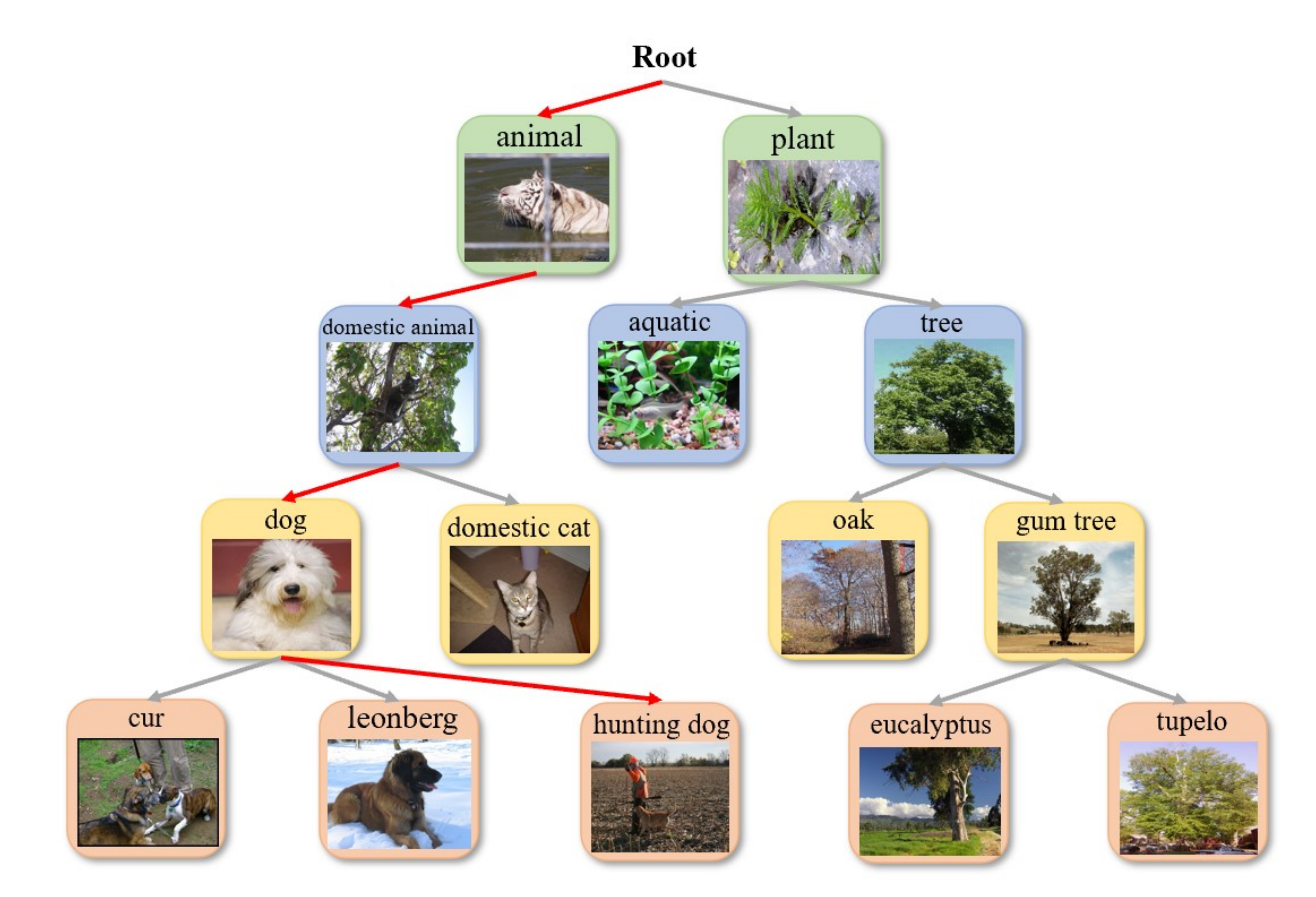}    
    \caption{Intuitive illustration of our proposed HGR-Net. Suppose the ground truth is \texttt{Hunting Dog}, then we can find the real-label path \texttt{Root} $\to$ \texttt{Animal} $\to$ \texttt{Domestic Animal} $\to$ \texttt{Dog} $\to$ \texttt{Hunting Dog}. Our goal is to efficiently leverage semantic hierarchical information to help better understand the visual-language pairs.}
    \label{fig:tissue}
\end{figure}

A few works of literature explored zero-shot image classification on ImageNet-21K. However, the performance has plateaued to a few percent Hit@1 performances on ImageNet-21K zero-shot classification benchmark (~\cite{frome2013devise,norouzi2013zero,long2017describing,wang2018zero}). We believe the key challenge is distinguishing among 21K highly fine-grained classes. These methods  represents class information by GloVe~\cite{pennington2014glove} or Skip-Gram~\cite{mikolov2013distributed} to align the vision-language relationships. However, these lower-dimensional features from GloVe or Skip-Gram are not representative enough to distinguish among 21K classes, especially since they may collapse for fine-grained classes. Besides, most existing works train a held-out classifier to categorize images of unseen classes with different initialization schemes. One used strategy is to initialize the classifier weights with semantic attributes~\cite{yu2018stacked,liu2020hyperbolic,xie2019attentive,skorokhodov2021class}, while another is to conduct fully-supervised training with generated unseen images. However, the trained MLP-like classifier is not representative enough to capture fine-grained differences to classify the image into a class with high confidence. 

To resolve the challenge of large-scale zero-shot image classification, we proposed a novel \emph{H}ierarchical \emph{G}raph knowledge \emph{R}epresentation network (denoted as HGR-Net). We explore the conceptual knowledge among classes to prompt the distinguishability. In Fig.~\ref{fig:tissue}, we state the intuition of our proposed method. Suppose the annotated image class label is \texttt{Hunting Dog}. The most straightforward way is to extract the semantic feature and train the classifier with cross-entropy. However, our experiments find that better leveraging hierarchical conceptual knowledge is important to learn discriminative text representation. We know the label as \texttt{Hunting Dog}, but all the labels from the root can also be regarded as the real label. We incorporate conceptual semantic knowledge to enhance the network representation.  

Moreover, inspired by the recent success of pre-trained models from large vision-language pairs such as CLIP~\cite{radford2021learning} and ALIGN~\cite{jia2021scaling}, we adopt a dynamic confidence-based classification scheme, which means we multiply a particular image feature with candidate text features and then select the most confident one as the predicted label. Unlike traditional softmax-based classifier, this setting is dynamic, and no need to train a particular classifier for each task. Besides, the confidence-based scheme can help truly evaluate the vision-language relationship understanding ability. For better semantic representation, we adopt Transformer~\cite{vaswani2017attention} as the feature extractor, and follow-up experiments show Transformer-based text encoder can significantly boost the classification performance. 

\noindent \textbf{Contributions.} We consider the most challenging large-scale zero-shot image classification task on ImageNet-21K and proposed a novel hierarchical graph representation network, HGR-Net, to model the visual-semantic relationship between seen and unseen classes. Incorporated with a confidence-based learning scheme and a Transformers to represent class semantic information, we show that HGR-Net achieved new state-of-the-art performance with significantly better results than baselines. We also conducted few-shot evaluations of HGR, and we found our method can learn very efficiently by accessing only one example per class. We also conducted extensive experiments on the variants of ImageNet-21K, and the results demonstrate the effectiveness of our HGR-Net. To better align with our problem, we also proposed novel matrices to reflect the conceptual learning ability of different models.

\section{Related Work}
\subsection{Zero-/Few-Shot Learning}        
Zero-Shot Learning (ZSL) is recognizing images of unseen categories. Our work is more related to semantic-based methods, which learn an alignment between different modalities (i.e., visual and semantic modalities) to facilitate  classification~\cite{yu2018stacked,liu2020hyperbolic,xie2019attentive,skorokhodov2021class}. CNZSL~\cite{skorokhodov2021class} proposed to map attributes into the visual space by normalization over classes. In contrast to~\cite{skorokhodov2021class}, we map both the semantic text and the images into a common space and calculate the confidence. Experimental studies are conducted to show that mapping to a common space achieves higher accuracy. We also explore the Few-Shot Learning (FSL) task, which focuses on classification with only accessing a few testing examples during training~\cite{sun2020meta,zhang2020deepemd,ye2021learning}. Unlike~\cite{wang2020generalizing} which defines the FSL task as extracting few training data from all classes, we took all images from seen classes and selected only a few samples from unseen classes during training. Our main goal here is to analyze how the performance differs from zero to one-shot.

\subsection{Large-Scale Graphical Zero-Shot Learning}
Graphical Neural Networks~\cite{kipf2016semi} are widely applied to formulate zero-shot learning, where each class is associated with a graph node, and a graph edge represents each inter-class relationship. For example, \cite {wang2018zero} trains a GNN based on the WordNet knowledge to generate classifiers for unseen classes. Similarly, \cite{kampffmeyer2019rethinking} uses fewer convolutional layers but one additional dense connection layer to propagate features towards distant nodes for the same graph. More recently, \cite{nayak2020zero} adopts a transformer graph convolutional network (TrGCN) for generating class representations. \cite{wang2021zero} leverages additional neighbor information in the graph with a contrastive objective. Unlike these methods, our method utilizes fruitful information of a hierarchical structure based on class confidence and thus grasps hierarchical relationships among classes to distinguish hard negatives. Besides, some works exploit graphical knowledge without explicitly training a GNN. For example, \cite{lu2015unsupervised} employs semantic vectors of the class names using multidimensional scaling (MDS) \cite{cox2008multidimensional} on the WordNet to learn a joint visual-semantic embedding for classification; \cite{liu2020hyperbolic} learns similarity between the image representation and the class representations in the hyperbolic space.

\subsection{Visual Representation Learning from Semantic Supervision}
Visual representation learning is a challenging task and has been widely studied with supervised or self-supervised methods. Considering semantic supervision from large-scale unlabeled data, learning visual representation from text representation \cite{radford2021learning} is a promising research topic with the benefit of large-scale visual and linguistic pairs collected from the Internet. These methods train a separate encoder for each modality (i.e., visual and language), allowing for extended to unseen classes for zero-shot learning. Upon these methods, \cite{cheng2021data} improves the data efficiency during training, \cite{jia2021scaling} enables learning from larger-scale noisy image-text pairs, \cite{zhou2021learning} optimizes the language prompts for better classifier generation. Our work adopts the pre-trained encoders of~\cite{radford2021learning} but tackles the problem of large-scale zero-shot classification from a candidate set of 22K classes instead of at most 1K as in \cite{radford2021learning}.

\section{Method}
\subsection{Problem Definition}
\noindent \textbf{Zero-shot learning.}
Let $\mathcal{C}$ denote the set of all classes. $\mathcal{C}_{s}$ and $\mathcal{C}_{u}$ to be the unseen and seen classes, respectively, where  $\mathcal{C}_{s} \cap \mathcal{C}_{u} =\emptyset$, and  $\mathcal{C}=\mathcal{C}_{s} \cup \mathcal{C}_{u}$. For each class $c_i\in \mathcal{C}$, a $d$-dimensional semantic representation vector $t(c_i)\in \mathbb{R}^{d}$ is provided. We denote the training set $\mathcal{D}_{tr}=\{(\mathbf{x}_i, c_i, t(c_i))\}_{i=1}^{N}$, where $\mathbf{x}_i$ is the $i$-th training image. In ZSL setting, given testing images $\mathbf{x}_{te}$, we aim at learning a mapping function $\mathbf{x}_{te}\to \mathcal{C}_u$. In a more challenging setting, dubbed as  generalized ZSL, we not only aim at classifying images from unseen categories but also seen categories, where we learn $\mathbf{x}_{te}\to \mathcal{C}_u \cup \mathcal{C}_s$ covering the entire prediction space.

\noindent \textbf{Semantic hierarchical structure.}
We assume access to a semantic Directed Acyclic Graph (DAG), $\mathcal{G}=(\mathcal{V}, \mathcal{E})$, where $\mathcal{V} = \mathcal{C} \cup \left\{R\right\}$ and $\mathcal{E} \subseteq\left\{(x, y) \mid(x, y) \in \mathcal{C}^{2}\right.$,  $\left.x \neq y\right\}$. Here the two-tuple $(x, y)$ represents an parenting relationship between $x$ and $y$, which means $y$ is a more abstract concept than $x$. Here we manually add a root node $R$ with a in-degree of 0 into $\mathcal{G}$. For simplicity, given any node $c_i$, we denote the ordered set of all its ancestors obtained by shortest path search from $R$ to $c_i$ as  $\mathcal{A}^{c_i} = \left\{ a(c_i)_{j} \right\}_{j=1}^{N^a_i} \subseteq \mathcal{C}$. Similarly, we denote the set of all  siblings of $c_i$ as $\mathcal{S}^{c_i} = \left\{ s({c_i})_{j} \right\}_{j=1}^{N^s_i} \subseteq \mathcal{C}$. Finally,  $d(c_i) \triangleq \lvert \mathcal{A}^{c_i} \rvert$ is  defined as depth of node $c_i$. 

\begin{figure*}[!t]
    \centering
    \includegraphics[width=1.0\textwidth]{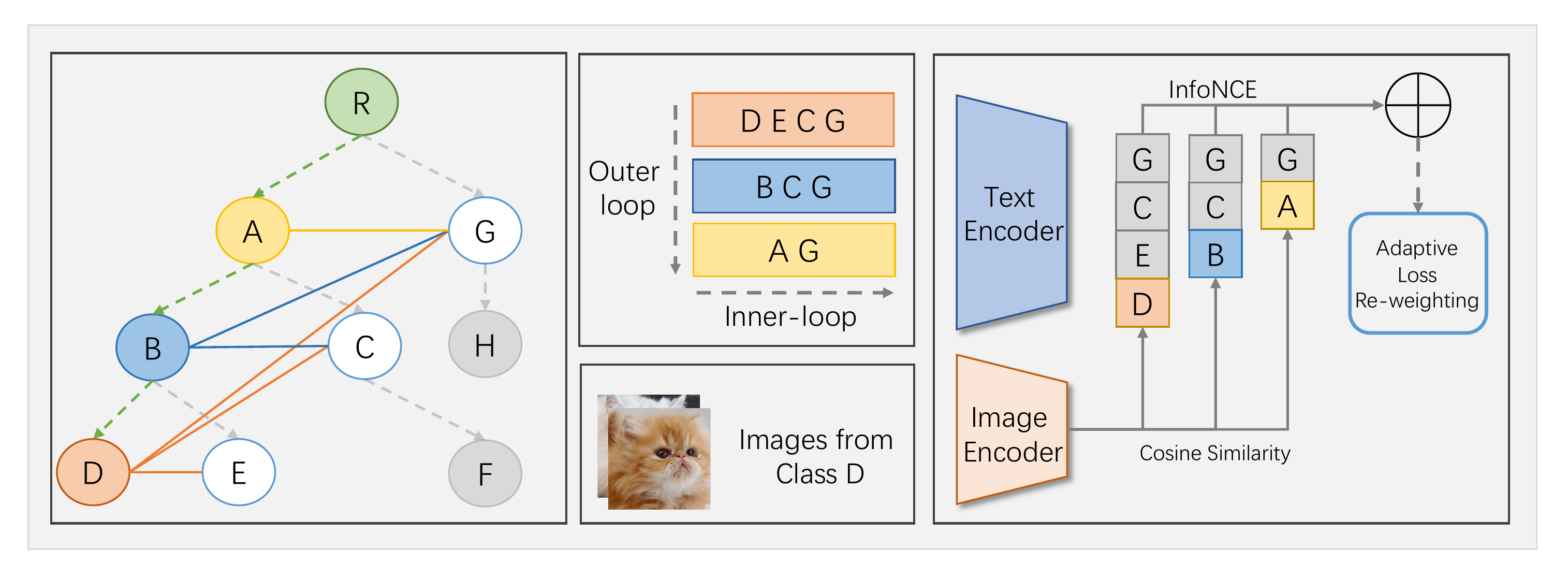}
    \caption{HGR-Net: Suppose the annotated single label is \texttt{D} and we can find the tracked label path $\texttt{R}\cdots\to \texttt{A}\to \texttt{B}\to \texttt{D}$ from the semantic graph extended from WordNet. We first set \texttt{D} as the positive anchor and contrast with negatives which are sampled siblings of its ancestors (i.e., $\{\texttt{E}, \texttt{C}, \texttt{G}\}$) layer by layer. Then we iterate to set the positive anchor to be controlled depth as $\texttt{B}, \texttt{A}$, which has layer-by-layer negatives $\{\texttt{C}, \texttt{G}\}$ and $\texttt{G}$, respectively. Finally, we use a memory-efficient adaptive  re-weighting strategy to fuse knowledge from different conceptual level.}
    \label{fig:main}
\end{figure*}

\subsection{HGR-Net: Large-Scale ZSL with Hierarchical Graph Representation Learning}
We mainly focus on zero-shot learning on the variants of ImageNet-21K, the current largest image classification dataset to our knowledge. Previous strategies~\cite{frome2013devise,norouzi2013zero,long2017describing,wang2018zero} adopt a $N$-way classification as the training task on  all the $N$ seen classes. However, we argue that this is problematic, especially in using a Transformer as the text encoder to obtain class semantic representations. First, when conducting $N$-way classification, all the classes except the single ground-truth are regarded as negative ones. Even though this helps build a well-performing classifier in a fully-supervised learning scenario, we argue that this is harmful to knowledge transfer from seen to unseen classes in ZSL. Second,  a batch of $N$ samples is fed to the text Transformer~\cite{vaswani2017attention} to obtain their corresponding text representations and to compute the class logits afterward. This strategy can be acceptable for datasets with a small number of classes. However, when the number of classes scales to tens of thousands, as in our case, it becomes formidable to implement the operations mentioned above. Therefore, we propose a memory-efficient hierarchical contrastive objective to learn transferable and discriminative representations for ZSL.
  
Intuitively, as illustrated in Fig.~\ref{fig:main}, suppose we have an image sample with annotated ground-truth label \texttt{D} according to ImageNet-21K. Then, we could find a shortest path $\texttt{R} -\cdots \to \texttt{A}\to \texttt{B}\to \texttt{D}$ to be the tracked true-label path $\mathcal{T}_\texttt{E}$. With our definition of the hierarchical structure, the true labels for the image sample are defined by all the nodes along this path through different levels of conceptions, from abstract to concrete in our case. Therefore, to better leverage this hierarchical semantic knowledge, we propose a hierarchical contrastive loss that conducts two levels of pre-defined degrees of bottom-up contrasting.

Specifically, for node \texttt{D} with a depth of $d(\texttt{D})$. In the outer-level loop, we iterate ground-truth labels of different levels, along the ancestor path $\mathcal{A}^{\texttt{D}}$, we traverse from itself bottom-up $\texttt{D} -\cdots \to \texttt{B}\to \texttt{A}$ until reaching one of its ancestors of a depth of $K d(\texttt{D})$, where $K$ is the outer ratio. In the inner-level loop, fixing the ground-truth label, we conduct InfoNCE loss~\cite{Aaron2018representation} layer by layer in a similar bottom-up strategy with an inner ratio $M$, (e.g., when fixing current ground truth node as \texttt{B} in Fig.~\ref{fig:main}, for inner loop we consider $\left\langle \texttt{B}, \texttt{C}\right\rangle, \left\langle \texttt{B}, \texttt{G}\right\rangle$). We provide more details in Alg.~\ref{alg:method}.

Formally, given an image of $\boldsymbol{x}$ from class $c_i$, we define its loss as:

\begin{equation}
    \mathcal{L_{\operatorname{cont}}} = \sum_{j=k_s}^{k_e}g(j, \mathcal{L}_j),\quad \mathcal{L}_{j} = \frac{1}{m_e-m_s+1}\sum_{l=m_s}^{m_e}\mathcal{L}_{j, l},
    \label{eq:l}
\end{equation}

\noindent where $g(\cdot, \cdot)$ is an adaptive attention layer to dynamically re-weight the importance of labels given different levels $j$, $j \in [k_s, k_e]$ and $l \in [m_s, m_e]$ are the outer-level and inner-level loop respectively. $k_s, k_e$ represents the start layer and the end layer for outer loop while $m_s, m_e$ are the start layer and the end layer for the inner loop. 


\begin{equation}
    \mathcal{L}_{j, l}=-\log \frac{\mathrm{pos}^j}
    {\mathrm{pos}^j + \mathrm{neg}^{j, l}},
    \label{eq:lj}
\end{equation}

\noindent where 
\begin{equation}
    \begin{aligned}
    \mathrm{pos^j} &= \exp \left(\operatorname{sim}\left(T(\boldsymbol{c}_{j}^{+}), V(\boldsymbol{x})\right)
    / \tau\right)
    \end{aligned}
    \label{eq:posneg1}
\end{equation}
\begin{equation}
    \begin{aligned}
        \qquad \mathrm{neg}^{j, l} &= \sum_{q=1}^{n_l} \exp \left(\mathrm{sim}\left(T(\boldsymbol{c}^{-}_{j, l, q}), V(\boldsymbol{x})\right) / \tau\right) 
    \end{aligned}
    \label{eq:posneg2}
\end{equation}

\noindent where, $\mathrm{sim}(\cdot)$ is the measure of similarity, $\tau$ is the temperature value. $V(\cdot)$ and $T(\cdot)$ are the visual and image encoders,  $\boldsymbol{c}_j^{+} = a(c_i)_{j}$ is the selected positive label on the tracked lable path at layer $l$. $\boldsymbol{c}_{j, l, q}^{-}$ is the $q$-$th$ sibling of the $j$-$th$ ground-truth at level $l$; see Alg.~\ref{alg:method}.

\begin{algorithm}[!htbp]
\caption{Hierarchical Graph Representation Net (HGR-Net)}\label{alg:method}
\begin{algorithmic} 
\Require Training set $\mathcal{D}_{\operatorname{tr}}$, text encoder $T$, visual encoder $V$, inner ratio $M$, outer ratio $K$, per layer sampling number threshold $\epsilon$, training label set $\mathcal{C}$, hierarchical graph $\mathcal{G}$
\State Sample a batch of data $\mathbf{X}$ from class $c_i$
\State Obtain its ancestor path $\mathcal{A}^{c_i}$
\State Set the outer loop range $k_s=K d(c_i), k_e= d(c_i)$ 
\For{$j = k_s, k_s+1, \dots, k_e$}
    \State Set the current ground-truth label $\boldsymbol{c}^{+}_{j}=a(c_i)_j$
    \State Prepare $\mathrm{pos^j}$ according to Eq. \ref{eq:posneg1}
    \State Set the inner-loop ranges $m_s = M d(\boldsymbol{c}^{+}), m_e = d(\boldsymbol{c}^{+})$
    \For{$l = m_s, m_s+1, \dots, m_e$}
        \State Prepare sibling set $\mathcal{S}^{c_j}$ 
        \State $n_l = \mathrm{max}(\epsilon, \lvert \mathcal{S}^{c_j} \rvert)$
        \State Sample $n_l$ negative sibling set $\left \{ \boldsymbol{c}_{j, l, q}^{-}\right \}_{q=1}^{n_l}$
        \State Prepare $\mathrm{neg^{j, l}}$ according to Eq.~\ref{eq:posneg2}
        \State Compute $\mathcal{L}_{j, l}$ according to Eq.~\ref{eq:lj}
    \EndFor
    \State Compute $\mathcal{L}_{j}$ according to Eq.~\ref{eq:l} right part
\EndFor
\State Compute $\mathcal{L}_{\mathrm{cont}}$ according to Eq.~\ref{eq:l} left part
  
\end{algorithmic}
\end{algorithm}

\section{Experiments}   

\subsection{Datasets and the Hierarchical Structure}
ImageNet~\cite{deng2009imagenet} is a widely used large-scale benchmark for ZSL organized according to the WordNet hierarchy~\cite{miller1995wordnet}, which can lead our model to learn the hierarchical relationship among classes. However, the original hierarchical structure is not a DAG (Directed Acyclic Graph), thus not suitable when implementing our method. Therefore, to make all of the classes fit into an appropriate location in the hierarchical DAG, we reconstruct the hierarchical structure by removing some classes from the original dataset, which contains seen classes from the ImageNet-1K and unseen classes from the ImageNet-21K (winter-2021 release), resulting a modified dataset ImageNet-21K-D (D for Directed Acyclic Graph).

It is worth noticing that although there are 12 layers in the reconstructed hierarchical tree in total, most nodes reside between $2^{\operatorname{nd}}$ and $6^{\operatorname{th}}$ layers. Our class-wise dataset split is based on GBU~\cite{gbu}, which provides new dataset splits for ImageNet-21K with 1K seen classes for training and the remaining 20, 841 classes as test split. Moreover, GBU~\cite{gbu} splits the unseen classes into three different levels, including "2-hop", "3-hops" and "All" based on WordNet hierarchy~\cite{miller1995wordnet}. More specifically, the "2-hops" unseen concepts are within 2-hops from the known concepts. After the modification above, the training is then conducted on the processed ImageNet-1K with seen 983 classes, while 17,295 unseen classes from the processed ImageNet-21K are for ZSL testing, and 1533 and 6898 classes for "2-hops" and "3-hops" respectively. Please note that there is no overlap between the seen and unseen classes. The remaining 983 seen classes make our training setting more difficult because our model gets exposed to fewer images than the original 1k seen classes. Please refer to the supplementary materials for more detailed descriptions of the dataset split and reconstruction procedure.

\subsection{Implementation Details}
We use a modified ResNet-50~\cite{he2016deep} from~\cite{radford2021learning} as the image encoder, which replaces the global average pooling layer with an attention mechanism, to obtain visual representation with feature dimensions of 1024. Text descriptions are encoded into tokens and bracketed with start tokens and end tokens based on byte pair encoding (BPE)~\cite{sennrich-etal-2016-neural} with the max length of 77. For text embedding, we use CLIP~\cite{radford2021learning} Transformer to extract semantic vectors with the same dimensions as feature representation. We obtain the logits with L2-normalized image and text features and calculate InfoNCE loss~\cite{Aaron2018representation} layer by layer with an adaptive re-weighting strategy. More specifically, a learnable parameter with a size equivalent to the depth of the hierarchical tree is used to adjust the weights adaptively. 

\noindent \textbf{Training details.} We use the AdamW optimizer~\cite{loshchilov2018decoupled} applied to all weights except the adaptive attention layer with a learning rate 3e-7. We use the SGD optimizer for the adaptive layer with a learning rate of 1e-4. When computing the matmul product of visual and text features, a learnable temperature parameter $\tau$ is initialized as 0.07 from~\cite{Bastiaan2018scale} to scale the logits and clips gradient norm of the parameters to prevent training instability. To accelerate training and avoid additional memory, mixed-precision~\cite{micikevicius2017mixed} is used, and the weights of the model are only transformed into float32 for optimization. Our proposed HGR model is implemented in PyTorch, and training and testing are conducted on a Tesla V100 GPU with a batch size of 256 and 512, respectively.

\subsection{Large-Scale ZSL Performance}


\textbf{Comparison approaches.}
We compare with the following approaches: 

\noindent -- \textbf{DeViSE~\cite{frome2013devise}} linearly maps visual information to the semantic word-embedding space. The transformation is learned using a hinge ranking loss.

\noindent -- \textbf{HZSL~\cite{liu2020hyperbolic}} learns similarity between the image representation and the class representations in the hyperbolic space. 

\noindent -- \textbf{SGCN~\cite{kampffmeyer2019rethinking}} uses an asymmetrical normalized graph Laplacian to learn the class representations.

\noindent -- \textbf{DGP~\cite{kampffmeyer2019rethinking}} separates adjacency matrix into ancestors and descendants and propagates knowledge in two phases with one additional dense connection layer based on the same graph as in GCNZ~\cite{wang2018zero}.

\noindent -- \textbf{CNZSL~\cite{skorokhodov2021class}} utilizes a simple but effective class normalization strategy to preserve variance during a forward pass.

\noindent -- \textbf{FREE~\cite{Chen2021free}} incorporates semantic-visual mapping into a unified generative model to address cross-dataset bias.

\noindent \textbf{Evaluation Protocols.}
We use the typical Top@K criterion, but we also introduce additional metrics. Since it could be more desirable to have a relatively general but correct prediction rather than a more specific but wrong prediction, the following three metrics evaluate a given model's ability to learn the hierarchical relationship between the ground truth and its general classes.

\begin{itemize}
    \item \textbf{Top-Overlap Ratio (TOR).} In this metric, we take a further step to also cover all the ancestor nodes of the ground truth class. More concretely, for an image $x_j$ from class $c_i$ of depth $q_{c_i}$, TOR is defined as:
    \begin{equation}
        TOR(x_j) = \frac{\lvert p_{x_j} \cap \{A_{c_i},c_i\} \rvert} { q_{c_i} }
    \end{equation}
    where $c_i$ is the corresponding class to image $x_j$. $A_{c_i}$ is the union of all the ancestors of class $c_i$ and $p_{x_j}$ is the predicted class of $x_j$. In other words, this metric consider the predicted class correct if it is an ancestor of the ground truth.
    
    \item \textbf{Point-Overlap Ratio (POR).}
    In this setting, we let the model predict labels layer by layer. POR is defined as:
    \begin{equation}
        POR(x_j) = \frac{\lvert P_{x_j} \cap P_{c_i} \rvert}{ q_{c_i} },
    \end{equation}
    where
        $P_{c_i} = \{c_{i_1},c_{i_2},c_{i_3},\cdots,c_{i_{q_{c_i}-1}},c_{i}\}$
    is the union of classes from the root to the ground truth through all the ancestors, and $P_{x_j}$ is the union of classes predicted by our model layer by layer. 
     $q_{c_i}$ is count of all the ancestors including the ground truth label, which is tantamount to the depth of node $c_i$. The intersection calculates the overlap between correct and predicted points for image $x_j$.
\end{itemize}

\begin{table}[!htbp]
    \centering
\begin{adjustbox}{width=0.8\linewidth,center}
\begin{tabular}{c|ccccc|cc}
\hline  \multirow{2}{*}{ \textbf{Method} } & \multicolumn{5}{c}{ \textbf{Hit@} $\mathbf{k}(\%)$} & \multirow{2}{*}{ \textbf{TOR} } & \multirow{2}{*}{ \textbf{POR}} \\ 
 & 1 & 2 & 5 & 10 & 20 & \\ 
\hline
 Devise~\cite{frome2013devise} & $1.0$ & $1.8$ & $3.0$ & $15$ & $23.8$ & -& -\\
 HZSL~\cite{liu2020hyperbolic} & $3.7$ & $5.9$ & $10.3$ & $13.0$ & $16.4$ & - & -\\
 SGCN(w2v)~\cite{kampffmeyer2019rethinking} & 2.79& 4.49 & 8.26 & 13.05 & 19.49 & 4.97 & 10.01 \\
 SGCN(Tr)~\cite{kampffmeyer2019rethinking} & 4.83 & 8.17 & 14.61 & 21.23 & 29.42 & 8.33 & 14.69 \\
 DGP(w2v)~\cite{kampffmeyer2019rethinking} & $3.00$ & $5.12$ & $9.49$ & $14.28$ & $20.55$ & 7.07 & 11.71\\
 DGP(Tr)~\cite{kampffmeyer2019rethinking} & 5.78 & 9.57 & 16.89 & 24.09 & 32.62 & 12.39 & 15.50 \\
 CNZSL(w2v)~\cite{skorokhodov2021class} & 1.94 & 3.17 & 5.88 & 9.12 & 13.73 & 3.93 & 4.03 \\
 CNZSL(Tr)~\cite{skorokhodov2021class} & 5.77 & 9.48 & 16.49 & 23.25 & 31.00 & 8.32 & 7.22 \\
 FREE(w2v)~\cite{Chen2021free} & 2.87 & 4.91 & 9.54 & 13.28 & 20.36 & 4.89 & 5.37 \\
 FREE(Tr)~\cite{Chen2021free} & 5.76 & 9.54 & 16.71 & 23.65 & 31.81 & 8.59 & 9.68 \\
\cline{2-8}
 CLIP~\cite{radford2021learning} & $15.22$ & $22.54$ & $33.43$ & $42.13$ & $50.93$ & $18.55$ & 14.68\\
 HGR-Net(Ours) & $\mathbf{16.39}$ & $\mathbf{24.19}$ & $\mathbf{35.66}$ & $\mathbf{44.68}$ & $\mathbf{53.71}$ & $\mathbf{18.90}$ & $\mathbf{16.19}$ \\
\hline
\end{tabular}
\end{adjustbox}
    \caption{Top@k accuracy, Top-Overlap Ratio (TOR), and Point-Overlap Ratio (POR) for different models on the ImageNet-21K-D only testing on unseen classes. Tr means text encoder is CLIP Transformer.}
    \label{tab:compare}
\end{table}

\textbf{Results analysis.}
Tab.~\ref{tab:compare} demonstrates the performance of different models on ImageNet-21K ZSL setting on Top@K and above-mentioned three hierarchical evaluation. 
Our proposed model outperforms SoTA methods in all metrics, including hierarchical measures, proving the ability to learn the hierarchical relationship between the ground truth and its ancestor classes. We also attach the performance on 2-hops and 3-hops in the supplementary.

\subsection{Ablation Studies}
\textbf{Different attributes.}
Conventional attribute-based ZSL methods use GloVe~\cite{pennington2014glove} or Skip-Gram~\cite{mikolov2013distributed} as text models, while CLIP~\cite{radford2021learning} utilizes prompts (i.e., text description) template: "a photo of a \texttt{[CLASS]}", and take advantage of Transformer to extract text feature. Blindly adding Transformer to some attribute-based methods like HZSL~\cite{liu2020hyperbolic} which utilizes unique techniques to improve their performance in the attribute setting result in unreliable results. Therefore, we conducted experiments comparing three selected methods with different attributes. The result in Tab.~\ref{tab:attr} shows that methods based on text embedding extracted by CLIP transformer outperform traditional attribute-based ones since the low dimension representations (500-D) from w2v~\cite{mikolov2013distributed} is not discriminative enough to distinguish unseen classes, while higher dimension (1024-D) text representations significantly boost classification performance. Our HGR-Net gained significant improvement by utilizing Transformer compared to the low dimension representation from w2v~\cite{mikolov2013distributed}.

\begin{table}[!htbp]
    \centering
\begin{adjustbox}{width=\linewidth,center}
\begin{tabular}{c|c|ccccc|cc}
\hline \multirow{2}{*}{ \textbf{Attributes} } & \multirow{2}{*}{ \textbf{Methods} } & \multicolumn{5}{c}{ \textbf{Hit@} $\mathbf{k}(\%)$} & \multirow{2}{*}{ \textbf{TOR} } & \multirow{2}{*}{ \textbf{POR}}\\
& & 1 & 2 & 5 & 10 & 20 & \\ 
\hline
\multirow{7}{*}{w2v}
& SGCN~\cite{kampffmeyer2019rethinking} & 2.79& 4.49 & 8.26 & 13.05 & 19.49 & 4.97 & 10.01 \\
& DGP(w/o)~\cite{kampffmeyer2019rethinking} & 2.90 & 4.86 & 8.91 & 13.67 & 20.18 & 3.96 & 11.49 \\
& DGP~\cite{kampffmeyer2019rethinking} & $3.00$ & $5.12$ & $9.49$ & $14.28$ & $20.55$ & 7.07 & 11.71\\
& CNZSL(w/o CN)~\cite{skorokhodov2021class} & 0.83 & 1.47 & 3.03 & 5.08 & 8.27 & 1.98 & 2.05\\
& CNZSL(w/o INIT)~\cite{skorokhodov2021class} & 1.84 & 3.13 & 6.08 & 9.47 & 14.13 & 3.04 & 4.05 \\
& CNZSL~\cite{skorokhodov2021class} & 1.94 & 3.17 & 5.88 & 9.12 & 13.73 & 3.93 & 4.03 \\
& FREE~\cite{Chen2021free} & 2.87 & 4.91 & 9.54 & 13.28 & 20.36 & 4.89 & 5.37 \\
& HGR-Net(Ours) & 2.35 & 3.69 & 7.03 & 11.46 & 18.27 & 4.38 & 5.76 \\
\hline
\multirow{7}{*}{Transformer(CLIP)}
& SGCN~\cite{kampffmeyer2019rethinking} & 4.83 & 8.17 & 14.61 & 21.23 & 29.42 & 8.33 & 14.69 \\
& DGP(w/o)~\cite{kampffmeyer2019rethinking} & 5.42 & 9.16 & 16.01 & 22.92 & 31.20 & 7.80 & 15.29 \\
& DGP~\cite{kampffmeyer2019rethinking} & 5.78 & 9.57 & 16.89 & 24.09 & 32.62 & 12.39 & 15.50 \\
& CNZSL(w/o CN)~\cite{skorokhodov2021class} & 1.91 & 3.45 & 6.74 & 10.55 & 15.51 & 3.19 & 3.43 \\
& CNZSL(w/o INIT)~\cite{skorokhodov2021class} & 5.65 & 9.33 & 16.24 & 22.88 & 30.63 & 8.32 & 7.03 \\
& CNZSL~\cite{skorokhodov2021class} & 5.77 & 9.48 & 16.49 & 23.25 & 31.00 & 7.97 & 7.22 \\
& FREE~\cite{Chen2021free} & 5.76 & 9.54 & 16.71 & 23.65 & 31.81 & 8.59 & 9.68 \\
& HGR-Net(Ours) & $\mathbf{16.39}$ & $\mathbf{24.19}$ & $\mathbf{35.66}$ & $\mathbf{44.68}$ & $\mathbf{53.71}$ & $\mathbf{18.95}$ & $\mathbf{16.19}$ \\
\hline
\end{tabular}
\end{adjustbox}
    \caption{Different attributes. DGP(w/o) means without separating adjacency matrix into ancestors and descendants, \texttt{CN} and \texttt{INIT} in CNZSL means class normalization and proper initialization respectively.}
    \label{tab:attr}
\end{table}

\begin{figure}[!t]
    \centering
    \includegraphics[width=0.9\textwidth]{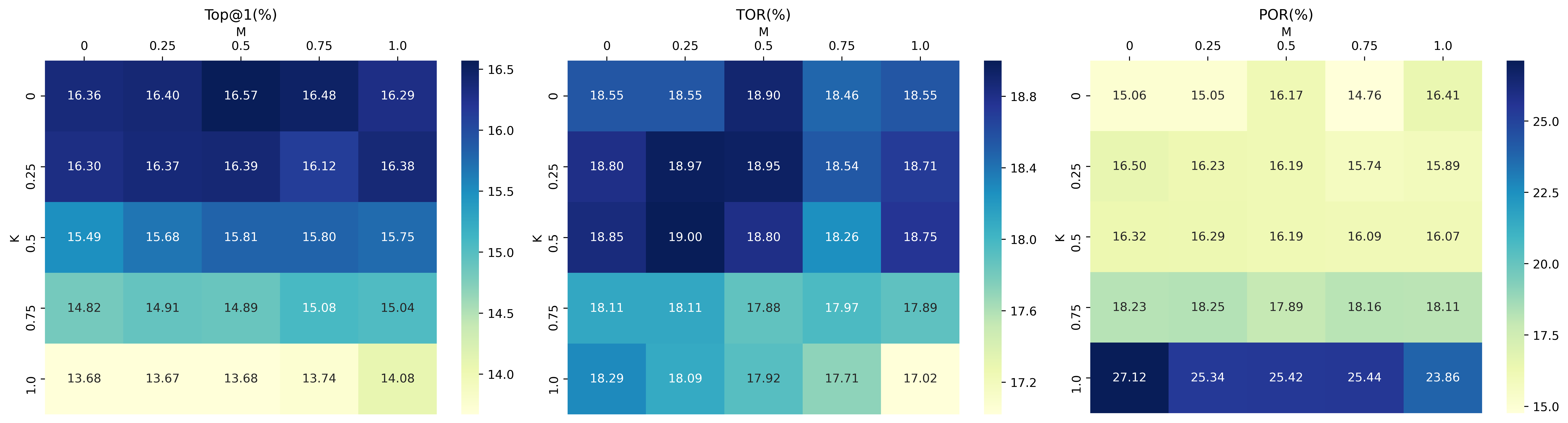}
    \caption{Different outer ratio (K) and inner ratio (M)}
    \label{fig:km_adaptive}
\end{figure}

\textbf{Different outer and inner ratio.}
Fig.~\ref{fig:km_adaptive} demonstrate the Top1, Top-Overlap Ratio (TOR) and Point-Overlap Ratio (POR) metrics of different \texttt{K} and \texttt{M}, where \texttt{K} and \texttt{M} $\in [0, 1]$. \texttt{K} and \texttt{M} are outer and inner ratio that determine how many samples is considered in the inner and outer loop respectively as earlier illustrated.

We explore different \texttt{K} and \texttt{M} in this setting and observe how performance differs under three evaluations. Please note that when \texttt{K} or \texttt{M} is 0.0, it means only the current node is involved in a loop. As \texttt{K} increases, the model is prone to obtain higher performance on hierarchical evaluation. An intuitive explanation is that more conceptual knowledge about ancestor nodes facilitates hierarchical learning relationships among classes.

\textbf{Different negative sampling strategies.}
We explore various sampling strategies for choosing negative samples and observe how they differ in performance. 
\emph{Random} randomly samples classes from all the classes.
\emph{TopM} samples neighbour nodes from $(q_{c_i}-M)$ to $q_{c_i}$ layers, where $q_{c_i}$ is the depth of inner anchor $c_i$, and we set M as 1.
\emph{Similarity} calculates the similarity of text features and chooses the top similar samples with the positive sample as hard negatives.
\emph{Sibling} samples sibling nodes of the target class. 
Tab.~\ref{tab:sampling} indicates that \emph{TopM} outperforms other sampling strategies. Therefore, we adopt the \emph{TopM} sampling strategy in the subsequent ablation studies.

\begin{table}[!htbp]
    \centering
\begin{adjustbox}{width=0.8\linewidth,center}
\begin{tabular}{c|ccccc|cc}
\hline
\multirow{2}{*}{ \textbf{Strategy} } & \multicolumn{5}{c}{ \textbf{Hit@} $\mathbf{k}(\%)$} & \multirow{2}{*}{ \textbf{TOR} } & \multirow{2}{*}{ \textbf{POR}}\\
& 1 & 2 & 5 & 10 & 20 & \\ 
\hline
Random & $15.72$ & $23.33$ & $34.69$ & $43.68$ & $52.73$ & $16.12$ & 13.04 \\
\hline
Sibling & $16.25$ & $23.95$ & $35.29$ & $44.16$ & $53.09$ & $17.91$ & $13.46$ \\
\hline
Similarity & $16.35$ & $24.04$ & $35.33$ & $44.17$ & $53.07$ & $18.60$ &$14.78$ \\
\hline
TopM(default) & $\mathbf{16.39}$ & $\mathbf{24.19}$ & $\mathbf{35.66}$ & $\mathbf{44.68}$ & $\mathbf{53.71}$ & $\mathbf{18.90}$ & $\mathbf{16.19}$ \\
\hline
\end{tabular}
\end{adjustbox}
    \caption{Analysis of sampling strategies}
    \label{tab:sampling}
\end{table}

\begin{table}[!htbp]
    \centering
    \begin{adjustbox}{width=0.8\linewidth,center}
\begin{tabular}{c|ccccc|cc}
\hline
\multirow{2}{*}{ \textbf{Weighting} } & \multicolumn{5}{c|}{ \textbf{Hit@} $\mathbf{k}(\%)$} & \multirow{2}{*}{ \textbf{TOR} } &  \multirow{2}{*}{ \textbf{POR}}\\
& 1 & 2 & 5 & 10 & 20 & \\ 
\hline
Adaptive(default) & $\mathbf{16.39}$ & $\mathbf{24.19}$ & $\mathbf{35.66}$ & $\mathbf{44.68}$ & $\mathbf{53.71}$ & $\mathbf{18.90}$ & $\mathbf{16.19}$ \\
\hline
Equal & 15.97 & 23.65 & 35.02 & 43.97 & 52.97 & 17.82 & 13.71 \\
\hline
Increasing $\uparrow$ & 15.85 & 23.50 & 34.85 & 43.83 & 52.83 & 17.80 & 13.81 \\
\hline
Decreasing $\downarrow$ & 16.08 & 23.77 & 35.16 & 44.10 & 53.09 & 17.84 & 13.59 \\
\hline
$\uparrow$ (non-linear) & 15.58 & 23.13 & 34.43 & 43.44 & 52.46 & 17.79 & 14.12 \\
\hline
$\downarrow$ (non-linear) & 16.19 & 23.89 & 35.26 & 44.18 & 53.13 & 17.87 & 13.47 \\
\hline

\end{tabular}
\end{adjustbox}
    \caption{Analysis of the weighting strategies when re-weighting in both inner and outer loop with K=0.25 and M=0.5.}
    \label{tab:weighting}
\end{table}

\textbf{Different weighting strategies.}
Orthogonal to negative sampling methods, we explore in this ablation the influence of different weighting strategies across the levels of the semantic hierarchy. The depth of the nodes in the hierarchical structure is not well-balanced, and the layers are not accessible for all objects. Therefore, it is necessary to focus on the importance of different layers. In this case, we experimented with 6 different weighting strategies and observed how they differ in multiple evaluations. As Tab~\ref{tab:weighting} shows, \emph{Increasing} gives more weights to deeper layers in a linear way and \emph{$\uparrow$ non-linear} is exponentially increasing weights to deeper layers. To balance the Top@K and hierarchical evaluations, the adaptive weighting method is proposed to obtain a comprehensive result. More specifically, \emph{Adaptive} uses a learnable parameter with a size equivalent to the depth of the hierarchical tree to adjust the weights adaptively. We attached the exact formulation of different weighting strategies in the supplementary.

\textbf{Experiment on ImageNet-21K-P~\cite{Tal2021masses}}
ImageNet-21K-P~\cite{Tal2021masses} is a pre-processed dataset from ImageNet21K by removing infrequent classes, reducing the number of total numbers by half but only removing only 13\% of the original images, which contains 12,358,688 images from 11,221 classes. We select the intersection of this dataset with our modified ImageNet21K dataset to ensure DAG structure consistency. The spit details (class and sample wise) are demonstrated in the supplementary. 

We show experimental results on ImageNet-21K-P comparing our method to different SoTA variants. Our model performs better in this smaller dataset compared to the original larger one in Tab.~\ref{tab:compare} and outstrips all the previous ZSL methods. We presented important results in Tab.~\ref{tab:ImageNet21KP} and we attached more results in the supplementary.

\begin{table}[!htbp]
    \centering
\begin{adjustbox}{width=0.8\linewidth,center}
\begin{tabular}{c|ccccc|cc}
\hline
\multirow{2}{*}{ \textbf{Models} } & \multicolumn{5}{c}{ \textbf{Hit@} $\mathbf{k}(\%)$} & \multirow{2}{*}{ \textbf{TOR} } & \multirow{2}{*}{ \textbf{POR}}\\
& 1 & 2 & 5 & 10 & 20 & \\
\hline
CNZSL(Tr w/o CN)~\cite{skorokhodov2021class} & 3.27 & 5.59 & 10.69 & 16.17 & 23.33 & 5.32 & 7.68\\ 
\hline
CNZSL(Tr w/o INIT)~\cite{skorokhodov2021class} & 7.90 & 12.77 & 21.40 & 29.50 & 38.63 & 11.23 &  12.56\\ 
\hline
CNZSL(Tr)~\cite{skorokhodov2021class} & $7.97$ & $12.81$ & $21.75$ & $29.92$ & $38.97$ & 11.50 &$12.62$ \\
\hline
\hline
FREE(Tr)~\cite{Chen2021free} & 8.15 & 12.90 & 21.37 & 30.29 & 40.62 & 11.82 & 13.34 \\
\hline
CLIP~\cite{radford2021learning} & $19.33$ & $28.07$ & $41.66$ & $53.77$ & $61.23$ & 20.08 &$20.27$ \\
\hline
HGR-Net(Ours) & \textbf{20.08} & \textbf{29.35} & \textbf{42.49} & \textbf{52.47} & \textbf{62.00} & \textbf{23.43} & \textbf{23.22} \\
\hline
\end{tabular}
\end{adjustbox}
    \caption{Result of ImageNet21K-P~\cite{Tal2021masses}. DGP(w/o)~\cite{kampffmeyer2019rethinking} means without separating adjacency matrix into ancestors and descendants, \texttt{CN} and \texttt{INIT} in CNZSL~\cite{skorokhodov2021class} means class normalization and proper initialization respectively, and Tr is Transformer of CLIP for short.}
    \label{tab:ImageNet21KP}
\end{table}
 
\begin{figure}[!th]
    \centering
    \includegraphics[width=0.9\textwidth]{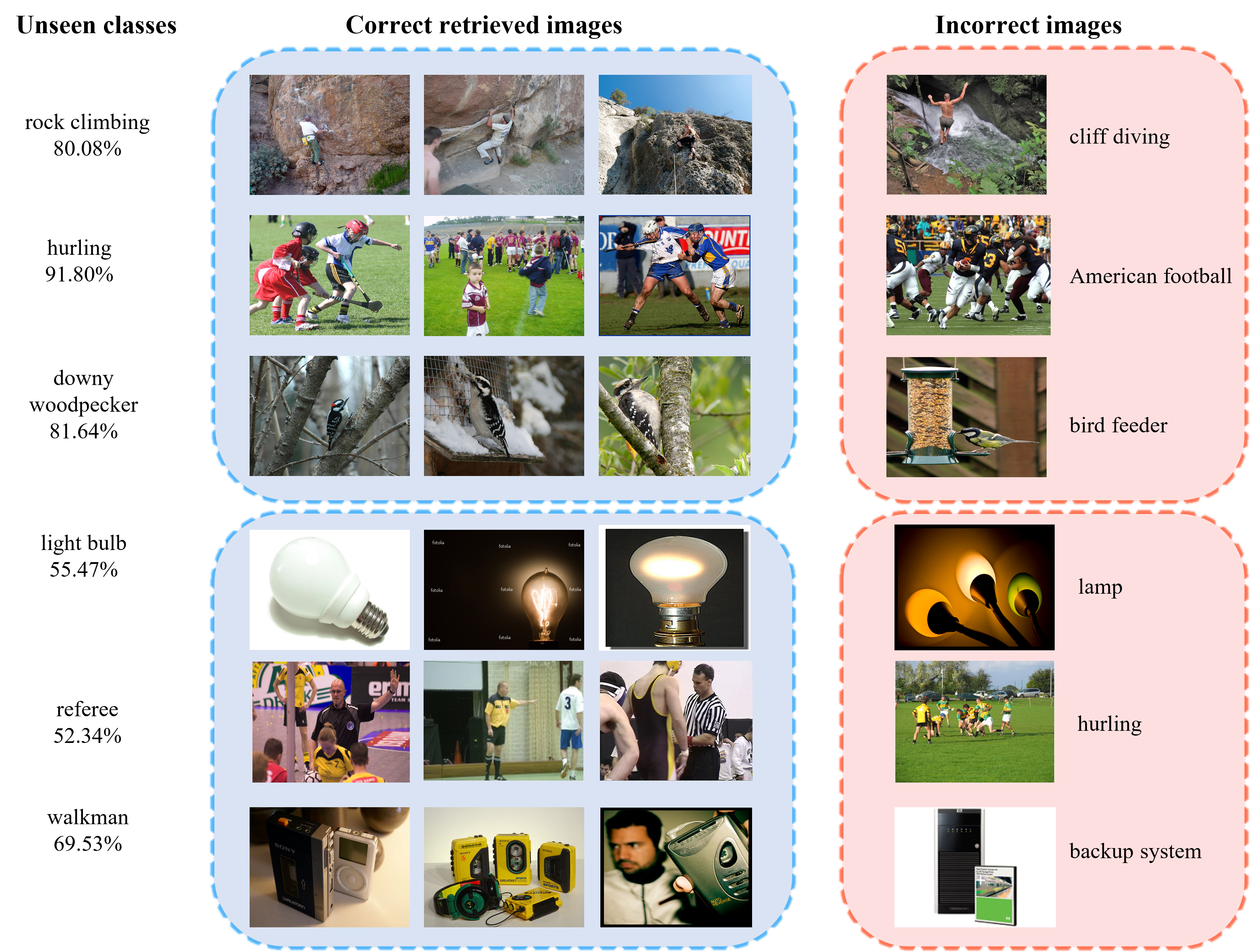}
    \caption{Zero-shot retrieved images. The first column represents unseen class names and corresponding confidence, the middle shows correct retrieval, and the last demonstrates incorrect images and their true labels.}
    \label{fig:retrieved}
\end{figure}

\subsection{Qualitative Results}

\begin{figure}[!htbp]
    \centering
    \includegraphics[width=1.0\textwidth]{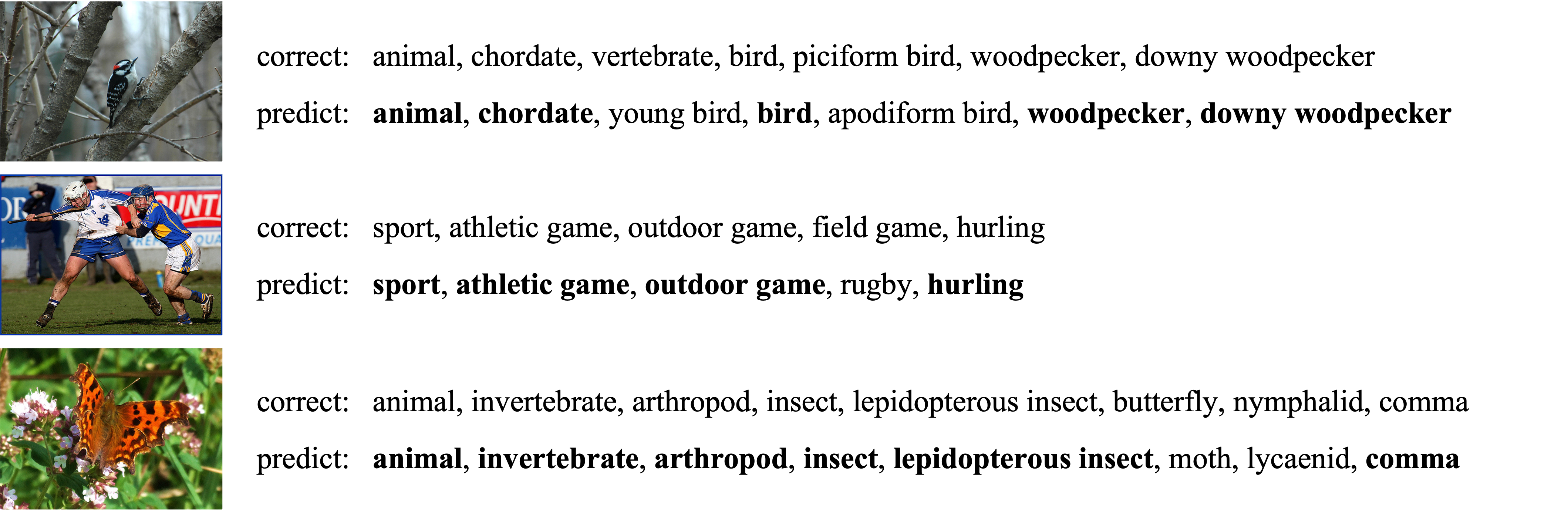}
    \caption{Predicted examples to show Point-Overlap. First row of each image is correct points from root to the ground truth and the second row show predicted points. The hit points are highlighted in bold.}
    \label{fig:point}
\end{figure}

Fig.~\ref{fig:retrieved} shows several retrieved images by implementing our model in the ZSL setting on ImageNet-21K-D. The task is to retrieve images from an unseen class with its semantic representation. Each row demonstrates three correct retrieved images and one incorrect image with its true label. Although our algorithm retrieves images from the wrong class, they are still visually similar to ground truth. For instance, the true label hurling and the wrong class American football belong to sports games, and images from both contain several athletes wearing helmets against a grass background. We also show some prediction examples in Fig.~\ref{fig:point} to present Point-Overlap results.

\subsection{Low-shot Classification on Large-Scale Dataset}

\begin{figure}[!h]
    \centering
    \includegraphics[width=\linewidth]{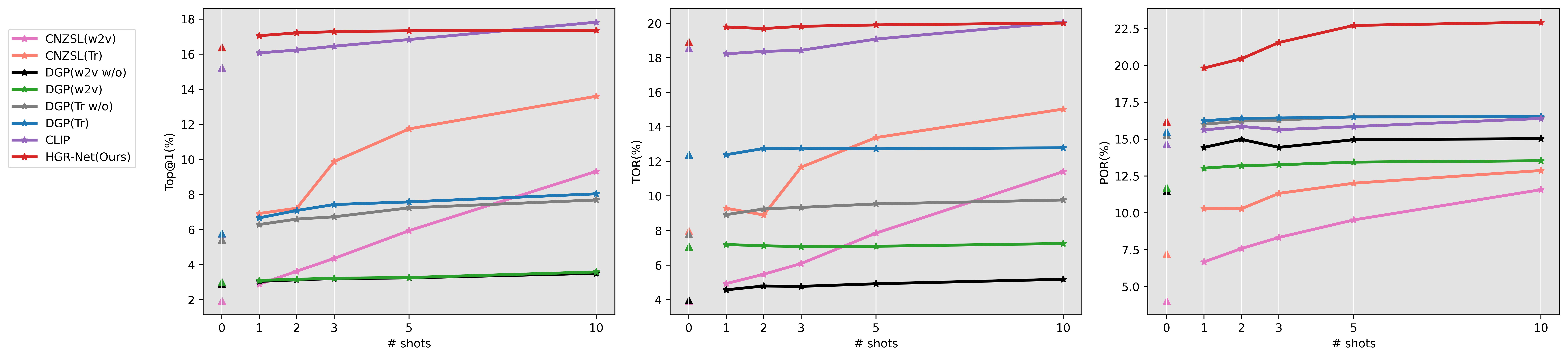}
    \caption{Few shots comparison. DGP(w/o)~\cite{kampffmeyer2019rethinking} means without separating adjacency matrix into ancestors and descendants, \texttt{CN} and \texttt{INIT} in CNZSL~\cite{skorokhodov2021class} means class normalization and proper initialization respectively, and Tr is Transformer of CLIP~\cite{radford2021learning} for short.}
    \label{fig:fewshot}
\end{figure}

Apart from zero-shot experiments being our primary goal in this paper, we also explore the effectiveness of our method in the low-shot setting compared to several baselines. Unlike pure few-shot learning, our support set comprises two parts. To be consistent with ZSL experiments, all the training samples of 983 seen classes are for low-shot training. For the 17, 295 unseen classes used in the ZSL setting, k-shots (1,2,3,5,10) images are randomly sampled for training in the low-shot setting, and the remaining images are used for testing. The main goal of this experiment is to show how much models could improve from zero to one shot and whether our proposed hierarchical-based method could generalize well in the low-shot scenario. Fig.~\ref{fig:fewshot} illustrated the few-shots results comparing our model to various SoTA methods. Although our approach gains trivial Top@k improvements from 1 to 10 shots, the jump from 0 to 1 shot is two times that from 1 to 10, proving that our model is an efficient learner.

\section{Conclusions}
This paper focuses on scaling-up visual recognition of unseen classes to tens of thousands of categories. We proposed a novel hierarchical graphic knowledge representation framework for confidence-based classification and demonstrated significantly better performance than baselines over Image-Net-21K-D and Image-Net-21K-P benchmarks, achieving new SOTA. We hope our work help ease future research of zero-shot learning and pave a steady way to understand large-scale visual-language relationships with limited data.

\noindent \subsubsection{Acknowledgments.} Research reported in this paper was supported by King Abdullah University of
Science and Technology (KAUST), BAS/1/1685-01-01.

\bibliographystyle{splncs04}
\bibliography{egbib}
     
\appendix

\newpage
\clearpage
The supplementary material provides: 
\begin{itemize}
  \item Section~\ref{section:dataset}: Additional details on dataset splits;
  \item Section~\ref{section:structure}: Description on reconstructed hierarchical structure;
  \item Section~\ref{section:implementation}: Additional training details;
  \item Section~\ref{section:performance}: Complementary experimental results on ImageNet-21K-P, ImageNet 2-hops and ImageNet 3-hops;
  \item Section~\ref{section:lowshot}: Experimental results on low-shot classification;
  \item Section~\ref{section:ablation}: Complementary strategy ablation study results and analysis.
  \item Section~\ref{section:add}: Additional ablations and comment.
\end{itemize}

\section{Dataset Description and Reconstructed Hierarchical Structure Details}\label{section:dataset}
To make all of the classes fit into an appropriate location in the hierarchical Directed Acyclic Graph (DAG), we remove \texttt{fa11misc} (Miscellaneous synsets not in the major subtrees in the ImageNet 2011 Fall Release) from the original hierarchical structure but add \texttt{food} and its sub-branches. And then, according to the reconstructed hierarchical structure, those irrelevant classes are also removed from the ImageNet-1K and ImageNet-21K (winter-2021 release), resulting in our ImageNet-21K-D dataset. The processed result is presented in Tab.~\ref{tab:dataset}.

\begin{table}[!htbp]
    \centering
\begin{adjustbox}{width=0.8\linewidth,center}
\begin{tabular}{c|ccccc}
\hline
{ \textbf{Class-wise Dataset} } & { \textbf{Train} } & { \textbf{Test} } & { \textbf{Train+Test} }  & { \textbf{2-hops} } & { \textbf{3-hops} }\\
\hline
Original & 1,000 & 20,841 & 21,841 & 1,549 & 7,860 \\
\hline
Processed & 983 & 17,295 & 18,278  & 1,533 & 6,898 \\
\hline
\end{tabular}
\end{adjustbox}
    \caption{Comparison between original ImageNet-21K (Original) and our ImageNet-21K-D (Processed).}
    \label{tab:dataset}
\end{table}

\begin{table}[!htbp]
	\centering
    \begin{adjustbox}{width=0.8\linewidth,center}
    \begin{tabular}{c|c|c|ccc}
    \hline
    { \textbf{Dataset} } & { \textbf{Description} } &{ \textbf{Setting} } & { \textbf{Train} } & { \textbf{Val} } & { \textbf{Test} } \\
    \hline
    \multirow{4}{*}{ImageNet-21K-D}
    &\multirow{2}{*}{\# of Classes}
    & seen & 983 & - & - \\
    && unseen & - & 17,295 & 17,295 \\ \cline{2-6}
    
    &\multirow{2}{*}{\# of Images}
    & seen & 1,259,303 & - & - \\
    && unseen & - & 792,510 & 11,337,589 \\
    \hline

    \multirow{4}{*}{ImageNet-21K-P}
    &\multirow{2}{*}{\# of Classes}
    & seen & 975 & - & - \\
    && unseen & - & 9,046 & 9,046 \\ \cline{2-6}
    
    &\multirow{2}{*}{\# of Images}
    & seen & 1,252,157 & - & - \\
    && unseen & - & 452,300 & 9,847,116 \\
    \hline
    \end{tabular}
    \end{adjustbox}
	\caption{ImageNet-21K-D and ImageNet-21K-P dataset split for ZSL.}
	\label{tab:zsl-split}
    \end{table}

First four lines in Tab.~\ref{tab:zsl-split} show our ImageNet-21K-D dataset splits. Please note that the official validation set, 50 images for each seen class, is used neither in ZSL training nor in ZSL validation. Our ZSL validation set is randomly sampled from unseen classes with at most 50 images per class, and all the images from unseen classes are used for ZSL testing.

ImageNet-21K-P~\cite{Tal2021masses} is a pre-processed dataset from ImageNet21K by removing infrequent classes, reducing the number of total numbers by half but only removing only 13\% of the original images. The original ImageNet-21K-P contains 12,358,688 images from 11,221 classes. After the above-mentioned pre-processing, class- and instance-wise splits are demonstrated in the rest four lines in Tab.~\ref{tab:zsl-split}.

\section{Hierarchical Structure}\label{section:structure}
Tab.~\ref{tab:examples} shows several examples of the reconstructed hierarchical tree. More general classes reside in the shallow layers, while the deeper layers contain more specific ones.

\begin{table}[!htbp]
    \centering
\begin{adjustbox}{width=0.8\linewidth,center}
\begin{tabular}{|c|c|}
\hline
{ \textbf{Layer} } & { \textbf{Example of Classes} } \\
\hline
1 & plant, sport, artifact, animal, person \\
\hline
2 & domestic animal, beach, painter \\
\hline
3 & wildflower, ice field, vertebrate \\
\hline
... & ... \\
\hline
12 & Atlantic bottlenose dolphin,  mouflon, Asian wild ox \\
\hline
\end{tabular}
\end{adjustbox}
    \caption{Example of classes in different layers. Noticed that we denote the root node as layer 0.}
    \label{tab:examples}
\end{table}

\begin{figure}[!htbp]
	\centering
	\includegraphics[width=0.7\textwidth]{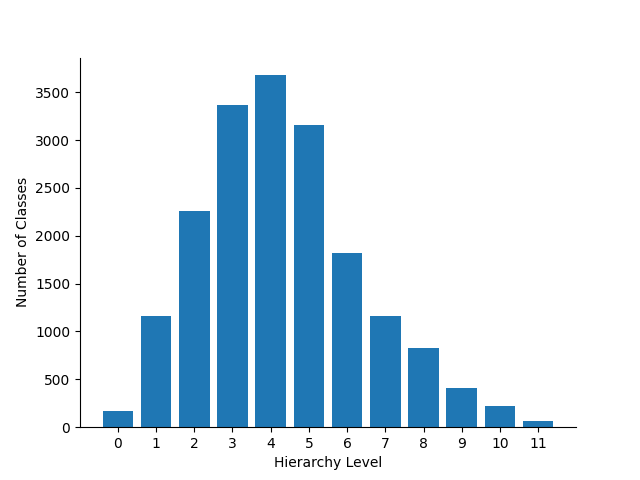}
	\caption{Distribution of classes in different layers on the ImageNet-21K-D dataset.}
	\label{fig:hierachy}
\end{figure}

Fig.~\ref{fig:hierachy} shows the imbalanced distribution of classes per layer in our reconstructed hierarchical tree. Although there are 12 layers in the reconstructed hierarchical tree, most nodes locate in $2^{th}-6^{th}$ layers.

\section{Implementation and Training Details}\label{section:implementation}

We choose ResNet-50~\cite{he2016deep} provided by CLIP~\cite{radford2021learning} as image encoder, which uses the ResNet-D improvements from~\cite{he2019bag} and the antialiased rect-2 blur pooling from~\cite{zhang2019making}, and replaces the global average pooling layer with an attention mechanism. The dimension of the extracted feature representation is 1024.
Moreover, our HGR-Net leverages two dimensions of hierarchical information. First, the ground truth class and all the ancestor nodes of this class tracing from the hierarchical tree are appended to form a list. We then set an outer ratio from this list to select part of the parent nodes as candidates for the outer loop. Similarly, within the outer loop, the inner ratio filters part of the parent nodes of the anchor node for the inner loop. In the inner loop, negative nodes of the outer-loop anchors are sampled through TopM sampling strategies. These negative classes contrast with the current layer's ground truth to guide the learning representation.
Text descriptions of negative classes, as well as the ground truth one, are encoded into tokens and bracketed with start tokens and end tokens based on byte pair encoding (BPE)~\cite{sennrich-etal-2016-neural} with the max length of 77. For text embedding, we use CLIP~\cite{radford2021learning} Transformer to extract semantic vectors with the same dimensions as feature representation. We obtain the logits with L2-normalized image and text features and calculate InfoNCE loss~\cite{Aaron2018representation} layer by layer with an adaptive re-weighting strategy. More specifically, a learnable parameter with a size equivalent to the depth of the hierarchical tree is used to adjust the weights adaptively in both the outer and inner loops.

We use the AdamW optimizer~\cite{loshchilov2018decoupled} applied to all weights except the adaptive attention layer with a learning rate 3e-7 and a default weight decay. The reason for choosing such a small learning rate is that we finetune CLIP~\cite{radford2021learning} with a hierarchical structure. Furthermore, a cosine scheduler is implemented to decay the learning rate for each step. In addition, we use the SGD optimizer separately for the adaptive layer with a learning rate of 1e-4. A learnable temperature parameter $\tau$ is initialized as 0.07 from~\cite{Bastiaan2018scale} to scale the logits, and gradient clipping is utilized to avoid training instability. Besides, to accelerate training and avoid additional memory, mixed-precision~\cite{micikevicius2017mixed} is used, and the weights of the model are only transformed into float32 for AdamW~\cite{loshchilov2018decoupled} optimization. The maximum number of sampled contrastive classes is set as 256. Training and testing are conducted on a Tesla V100 GPU with a batch size of 256 and 512, respectively.

\section{Performance Comparison}\label{section:performance}
Here we show complementary comparisons on more variances of the ImageNet dataset. We first show the results on the ImageNet-21K-P dataset~\cite{Tal2021masses}. The results show that our method achieved significantly better performance compared with baselines.

\begin{table}[!htbp]
    \centering
\begin{adjustbox}{width=0.8\linewidth,center}
\begin{tabular}{c|ccccc|cc}
\hline
\multirow{2}{*}{ \textbf{Models} } & \multicolumn{5}{c}{ \textbf{Hit@} $\mathbf{k}(\%)$} & \multirow{2}{*}{ \textbf{TOR} } & \multirow{2}{*}{ \textbf{POR}}\\
& 1 & 2 & 5 & 10 & 20 & \\
\hline
SGCN(w2v)~\cite{kampffmeyer2019rethinking} & 3.70 & 6.39 & 12.00 & 17.84 & 25.16 & 6.26 & 11.90 \\
\hline
DGP(w2v w/o)~\cite{kampffmeyer2019rethinking} & 3.88 & 6.62 & 11.85 & 17.54 & 25.14 & 5.11 & 12.04 \\
\hline
DGP(w2v)~\cite{kampffmeyer2019rethinking} & 4.01 & 6.72 & 12.10 & 17.93 & 25.77 & 8.14 & 13.30 \\ 
\hline
SGCN(Tr)~\cite{kampffmeyer2019rethinking} & 6.41 & 10.75 & 18.76 & 27.06 & 36.71 & 10.61 & 16.44\\
\hline
DGP(Tr w/o)~\cite{kampffmeyer2019rethinking} & 7.09 & 11.67 & 20.32 & 28.53 & 38.43 & 9.57 & 18.58\\
\hline
DGP(Tr)~\cite{kampffmeyer2019rethinking} & 7.20 & 11.95 & 20.98 & 29.81 & 39.62 & 14.59 & 19.33\\ 
\hline
CNZSL(w2v w/o CN)~\cite{skorokhodov2021class} & 1.25 & 2.22 & 4.54 & 7.56 & 12.21 & 3.04 & 5.17\\ 
\hline
CNZSL(w2v w/o INIT)~\cite{skorokhodov2021class} & 2.58 & 4.27 & 7.94 & 12.27 & 18.36 & 4.96 & 6.78\\ 
\hline
CNZSL(w2v)~\cite{skorokhodov2021class} & 2.58 & 4.27 & 8.01 & 12.44 & 18.58 & 3.88 & 6.58 \\
\hline
CNZSL(Tr w/o CN)~\cite{skorokhodov2021class} & 3.27 & 5.59 & 10.69 & 16.17 & 23.33 & 5.32 & 7.68\\ 
\hline
CNZSL(Tr w/o INIT)~\cite{skorokhodov2021class} & 7.90 & 12.77 & 21.40 & 29.50 & 38.63 & 11.23 &  12.56\\ 
\hline
CNZSL(Tr)~\cite{skorokhodov2021class} & $7.97$ & $12.81$ & $21.75$ & $29.92$ & $38.97$ & 11.50 &$12.62$ \\
\hline
FREE(w2v)~\cite{Chen2021free} & 3.95 & 6.32 & 11.85 & 16.57 & 24.93 & 5.76 & 8.31 \\
\hline
FREE(Tr)~\cite{Chen2021free} & 8.15 & 12.90 & 21.37 & 30.29 & 40.62 & 11.82 & 13.34 \\
\hline
CLIP~\cite{radford2021learning} & $19.33$ & $28.07$ & $41.66$ & $53.77$ & $61.23$ & 20.08 &$20.27$ \\
\hline
HGR-Net(Ours) & \textbf{20.08} & \textbf{29.35} & \textbf{42.49} & \textbf{52.47} & \textbf{62.00} & \textbf{23.43} & \textbf{23.22} \\
\hline
\end{tabular}
\end{adjustbox}
    \caption{Result of ImageNet21K-P~\cite{Tal2021masses}. DGP(w/o)~\cite{kampffmeyer2019rethinking} means without separating adjacency matrix into ancestors and descendants, \texttt{CN} and \texttt{INIT} in CNZSL~\cite{skorokhodov2021class} means class normalization and proper initialization respectively, and Tr is Transformer of CLIP for short.}
    \label{tab:ImageNet21KP}
\end{table}

We also conduct the performance comparison on two smaller ImageNet-21K variants, i.e., 2-hops, 3-hops. Tab.~\ref{tab:hops} proves the effectiveness of our method on smaller datasets. The performance drops for all models on "3-hops" test set than "2-hops" since seen classes are similar to unseen classes on "2-hops", while distant from unseen classes on "3-hops". However, our method still outperforms others when unseen classes are dominant in number and share less resemblance with seen classes, which proves efficiency in knowledge transferring.

\begin{table}[!htbp]
    \centering
\begin{adjustbox}{width=0.8\linewidth,center}
\begin{tabular}{c|c|ccccc|cc}
\hline \multirow{2}{*}{ \textbf{Test Set} } &  \multirow{2}{*}{ \textbf{Method} } & \multicolumn{5}{c}{ \textbf{Hit@} $\mathbf{k}(\%)$} & \multirow{2}{*}{ \textbf{TOR} } & \multirow{2}{*}{ \textbf{POR}} \\ 
& & 1 & 2 & 5 & 10 & 20 & \\ 
\hline
\multirow{7}{*}{2-hops}
& SGCN(w2v)~\cite{kampffmeyer2019rethinking} & 24.47 & 37.84 & 57.22 & 69.68 & 79.41 & 32.76 & 36.38 \\
& SGCN(Tr)~\cite{kampffmeyer2019rethinking} & 28.19 & 42.57 & 61.69 & 72.89 & 81.48 & 37.74 & 38.10 \\
& DGP(w2v)~\cite{kampffmeyer2019rethinking} & 24.57 & 37.67 & 56.88 & 69.60 & 79.17 & 34.94 & 37.04 \\
& DGP(Tr)~\cite{kampffmeyer2019rethinking} & 29.47 & 43.87 & 62.79 & 74.65 & 83.14 & 39.98 & 41.25 \\
& CNZSL(w2v)~\cite{skorokhodov2021class} & 11.99 & 19.11 & 32.46 & 44.31 & 56.40 & 17.37 & 16.70 \\
& CNZSL(Tr)~\cite{skorokhodov2021class} & 27.17 & 40.20 & 57.45 & 67.86 & 76.08 & 32.27 & 24.29 \\
& CLIP~\cite{radford2021learning} & 35.24 & 48.51 & 65.01 & 74.61 & 81.96 & 39.34 & 41.99 \\
& HGR-Net(Ours) & $\mathbf{36.11}$ & $\mathbf{49.46}$ & $\mathbf{65.90}$ & $\mathbf{75.69}$ & $\mathbf{82.98}$ & $\mathbf{40.87}$ & $\mathbf{42.63}$ \\
\hline
\multirow{7}{*}{3-hops}
& SGCN(w2v)~\cite{kampffmeyer2019rethinking} & 4.87 & 8.73 & 17.21 & 25.46 & 35.55 & 8.20 & 24.05 \\
& SGCN(Tr)~\cite{kampffmeyer2019rethinking} & 8.31 & 13.44 & 23.59 & 33.51 & 44.57 & 13.50 & 28.42 \\
& DGP(w2v)~\cite{kampffmeyer2019rethinking} & 4.95 & 8.81 & 16.91 & 25.64 & 36.14 & 10.40 & 26.85 \\
& DGP(Tr)~\cite{kampffmeyer2019rethinking} & 9.81 & 15.85 & 26.78 & 36.95 & 48.11 & 18.2 & 30.13 \\
& CNZSL(w2v)~\cite{skorokhodov2021class} & 3.71 & 6.11 & 11.31 & 17.21 & 24.95 & 7.14 & 16.08 \\
& CNZSL(Tr)~\cite{skorokhodov2021class} & 10.31 & 16.37 & 27.08 & 36.60 & 46.47 & 14.57 & 21.97 \\
& CLIP~\cite{radford2021learning} & 22.46 & 31.58 & 44.49 & 54.27 & 63.57 & 24.93 & 32.22 \\
& HGR-Net(Ours) & $\mathbf{23.23}$ & $\mathbf{32.53}$ & $\mathbf{45.74}$ & $\mathbf{55.70}$ & $\mathbf{65.05}$ & $\mathbf{26.54}$ & $\mathbf{32.44}$ \\
\hline
\end{tabular}
\end{adjustbox}
    \caption{Performance comparison among SoTA on 2-hops and 3-hops. Tr means text encoder is CLIP Transformer.}
    \label{tab:hops}
\end{table}
  
\section{Low-Shot Classification}\label{section:lowshot}
 
\begin{figure}[!htbp]
	\centering
	\includegraphics[width=\textwidth]{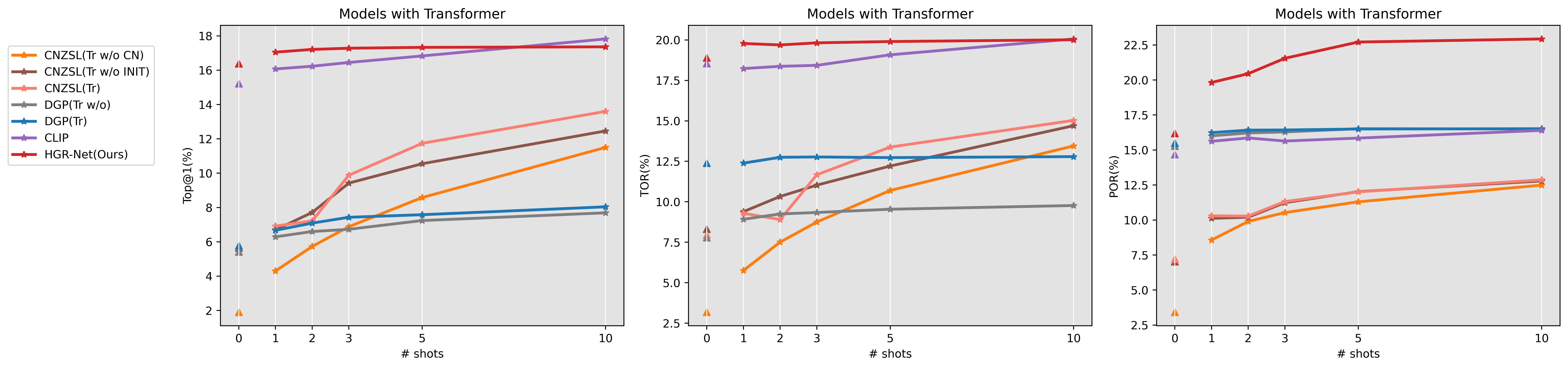}
	\caption{Top@1, Top-Overlap Ratio (TOR) and Point-Overlap Ratio (POR) results among different Transformer of ClIP~\cite{radford2021learning} based methods. DGP(w/o)~\cite{kampffmeyer2019rethinking} means without separating adjacency matrix into ancestors and descendants and Tr means the text encoder is Transformer of ClIP~\cite{radford2021learning}.}
	\label{fig:few_shot_tr}
\end{figure}

We have presented the performance in the main paper, but we select all the methods with Transformer of ClIP~\cite{radford2021learning} in an independent graph to make it more clear as Fig.~\ref{fig:few_shot_tr} shows.


\section{Complementary Ablation Study of Weighting Strategies}\label{section:ablation}

\begin{figure}[!t]
    \centering
    \includegraphics[width=\textwidth]{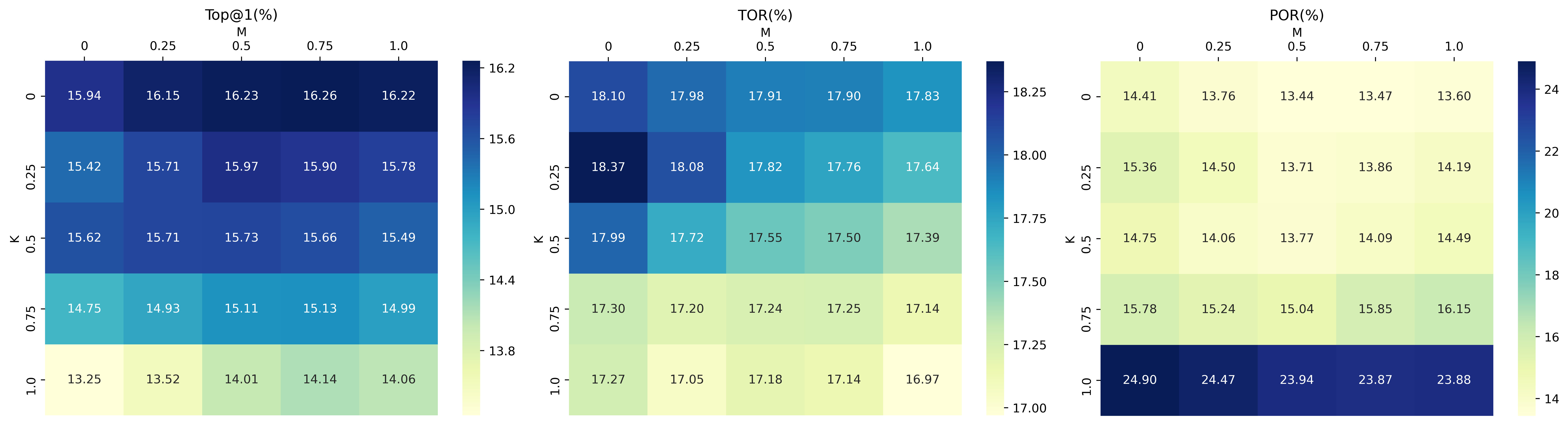}
    \caption{Different outer ratio (K) and inner ratio (M) with weighting strategy \emph{Equal}}
    \label{fig:km_equal}
\end{figure}

\begin{figure}[!t]
    \centering
    \includegraphics[width=\textwidth]{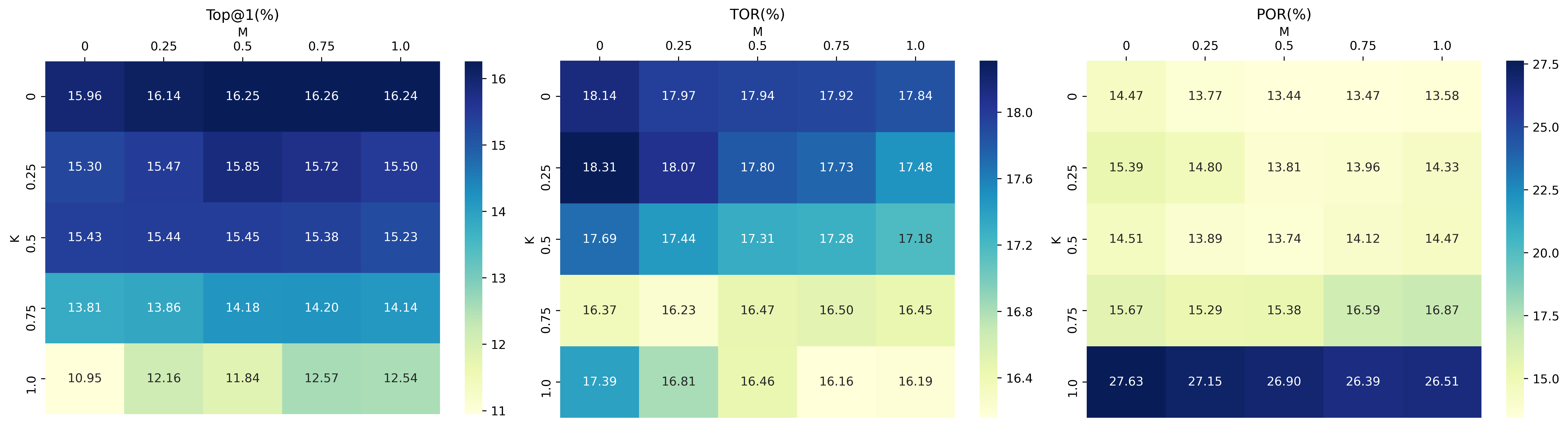}
    \caption{Different outer ratio (K) and inner ratio (M) with weighting strategy \emph{Increasing}}
    \label{fig:km_increasing}
\end{figure}

\begin{figure}[!t]
    \centering
    \includegraphics[width=\textwidth]{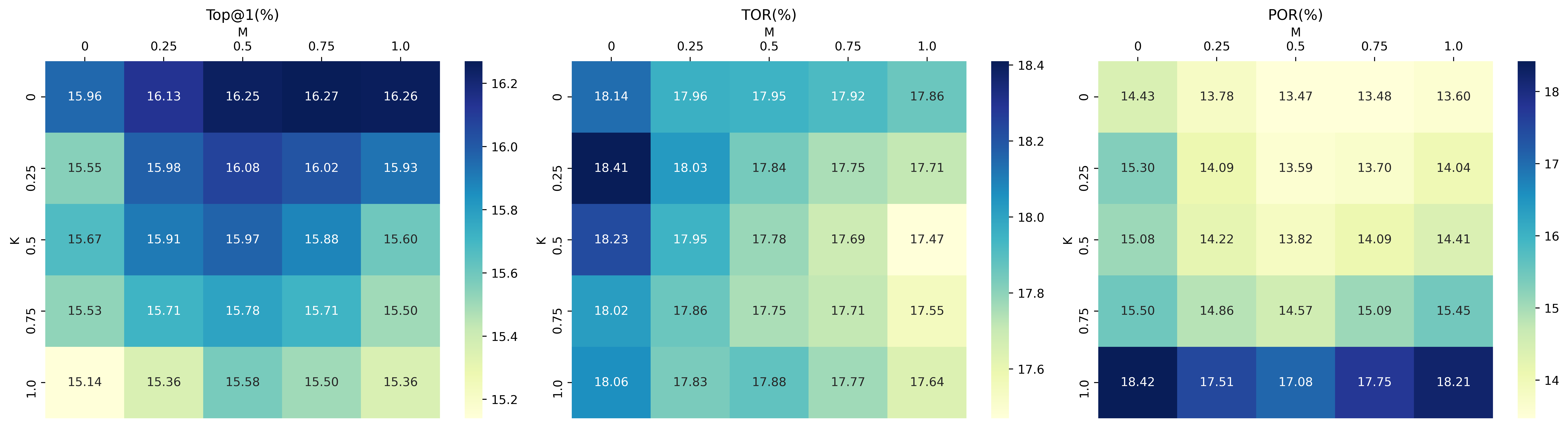}
    \caption{Different outer ratio (K) and inner ratio (M) with weighting strategy \emph{Decreasing}}
    \label{fig:km_decreasing}
\end{figure}

\begin{figure}[!t]
    \centering
    \includegraphics[width=\textwidth]{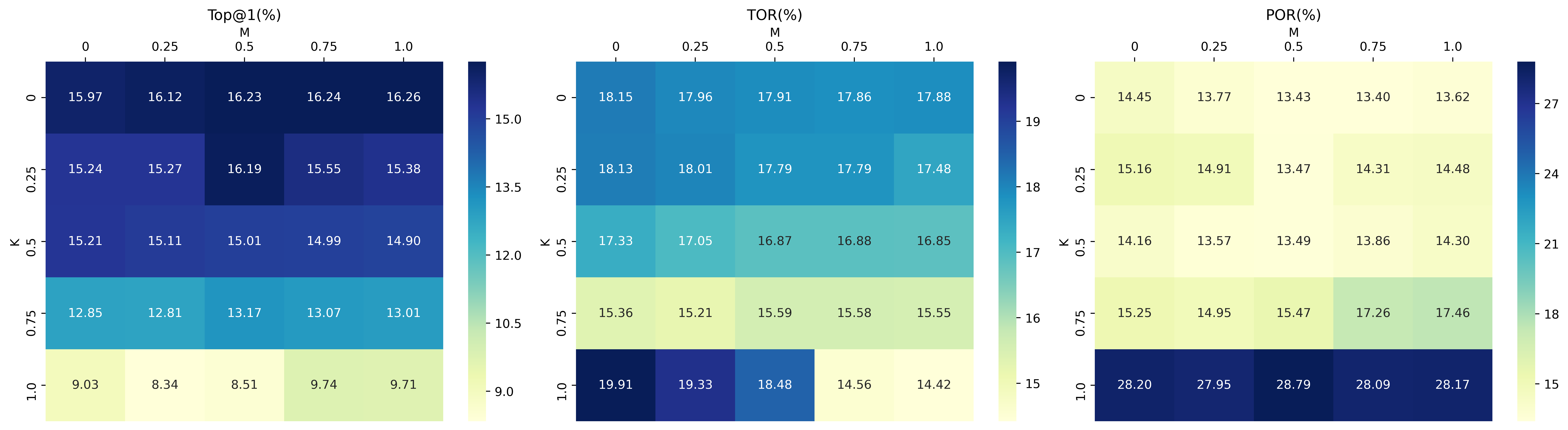}
    \caption{Different outer ratio (K) and inner ratio (M) with weighting strategy \emph{$\uparrow$ non-linear}}
    \label{fig:km_nl_increasing}
\end{figure}

\begin{figure}[!t]
    \centering
    \includegraphics[width=\textwidth]{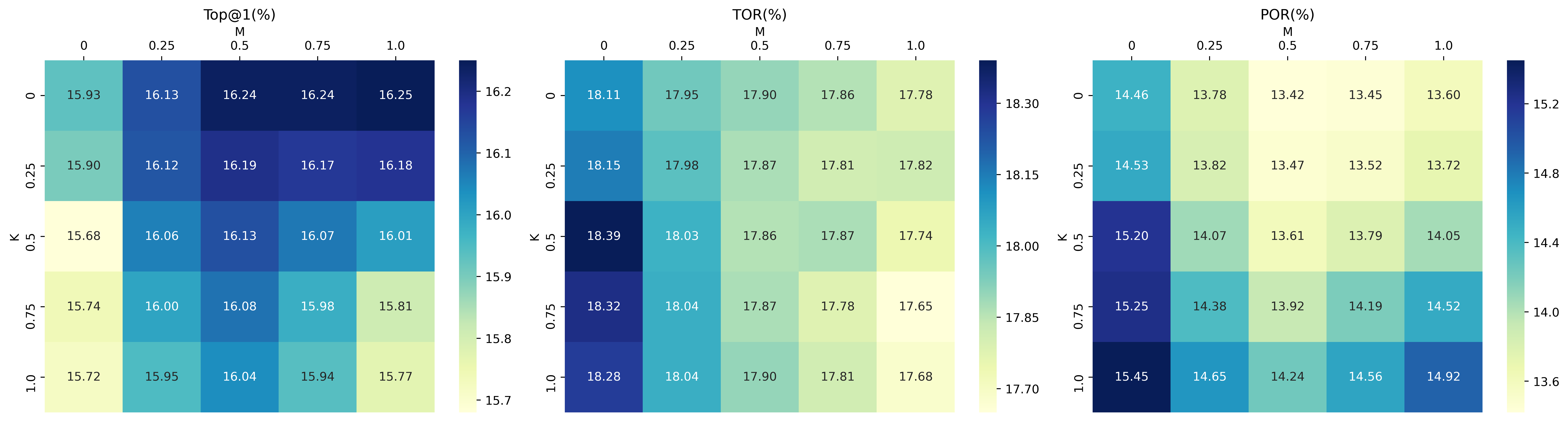}
    \caption{Different outer ratio (K) and inner ratio (M) with weighting strategy \emph{$\downarrow$ non-linear}}
    \label{fig:km_nl_decreasing}
\end{figure}

Fig.~\ref{fig:km_equal}~\ref{fig:km_increasing}~\ref{fig:km_decreasing}~\ref{fig:km_nl_increasing}~\ref{fig:km_nl_decreasing} show Top@1, Top-Overlap Ratio (TOR) and Point-Overlap Ratio (POR) evaluation among \emph{Equal}, \emph{Increasing}, \emph{Decreasing}, \emph{$\uparrow$ non-linear} and \emph{$\downarrow$ non-linear} weighting strategies in different outer ratio (\texttt{K}) and inner ratio (\texttt{M}).

Based on extensive experiments, the \emph{adaptive} weighting strategy with a learnable parameter obtained the best performance. The observed result of learned weights firstly decreases and then increases as the depth goes deeper, which can be deemed as inversely proportional to the number of classes in each layer. Therefore, the \emph{Adaptive} weighting can be simplified as: let $N_j$ be the total number of classes in each layer $j$, and $n$ be the number of layers, the weight for layer $j$ is defined as $\frac {\frac {1} {N_j}} {\sum_{i=0}^n \frac 1 {N_i}}$, in order to simplify the optimization and to achieve a stable result. The imbalanced number of classes can explain this result in different layers as Fig.~\ref{fig:hierachy} and the model learns to activate layers with fewer classes more frequently to balance classes among different hierarchies.

\section{Additional Ablations and Comment}\label{section:add}
\paragraph{Loss ablations.} We found it could be interesting to consider more ablations on our loss design, e.g., we consider two variants that both contain one single loop. The first one only considers the inner loss regarding the ground truth as the real label, while the second traces the ancestors of the ground truth as real labels and searches negative for each one. Experimental results demonstrate the performance drops by 5.61\% and 1.40\% respectively (see Table.~\ref{tab:compare} top part) and prove our outer-inner loops are necessary.

\paragraph{Experimental comparison fairness.}\label{exp_fair} Our model and the baseline DGP~\cite{kampffmeyer2019rethinking} both introduced the hierarchical semantic knowledge differently, but we demonstrated significantly better performance, which shows that a better-designed hierarchical graph can be critical to achieving good performance. The standard version of other baselines does not introduce a hierarchical graph in the training step. So we directly add HGR to the baseline CNZSL~\cite{skorokhodov2021class}, dubbed CNZSL+HGR. As Table~\ref{tab:compare} shows, leveraging our HGR  can increase the CNZSL~\cite{skorokhodov2021class} performance by 3.18\% on average. We suggest it doesn't improve much because CNZSL~\cite{skorokhodov2021class} has a shallower architecture than CLIP~\cite{radford2021learning}  and builds on top visual features instead of raw images.
Besides, our method HGR-Net still shows significantly better performance compared with CNZSL+HGR.

\begin{table}[!tb]
    \centering
\begin{adjustbox}{width=1.0\linewidth,center}
\begin{tabular}{l|ccccc|cc}
\hline  \multirow{2}{*}{ \textbf{Method} } & \multicolumn{5}{c}{ \textbf{Hit@} $\mathbf{k}(\%)$} & \multirow{2}{*}{ \textbf{TOR} } & \multirow{2}{*}{ \textbf{POR}} \\ 
 & 1 & 2 & 5 & 10 & 20 & \\ 
 \hline
 \textbf{HGR-Net(Ours)} & $\mathbf{16.39}$ & $\mathbf{24.19}$ & $\mathbf{35.66}$ & $\mathbf{44.68}$ & $\mathbf{53.71}$ & $\mathbf{18.90}$ & $\mathbf{16.19}$ \\
 $\quad$- outer loss & $15.47^{\textcolor{red}{-0.92}}$ & $22.80^{\textcolor{red}{-1.49}}$ & $33.65^{\textcolor{red}{-2.01}}$ & $42.23^{\textcolor{red}{-2.45}}$ & $50.97^{\textcolor{red}{-2.74}}$ & $17.00^{\textcolor{red}{-1.90}}$  & $15.80^{\textcolor{red}{-0.39}}$ \\
 $\quad$- inner loss & $16.16^{\textcolor{red}{-0.23}}$ & $24.00^{\textcolor{red}{-0.19}}$ & $35.47^{\textcolor{red}{-0.19}}$  & $44.38^{\textcolor{red}{-0.30}}$ & $53.46^{\textcolor{red}{-0.25}}$ & $18.49^{\textcolor{red}{-0.41}}$  & $16.09^{\textcolor{red}{-0.10}}$ \\
\hline
CNZSL(w2v) & 1.94 & 3.17 & 5.88 & 9.12 & 13.73 & 3.93 & 4.03 \\
$\quad$+HGR & $1.96^{\textcolor{blue}{+0.02}}$ & $3.20^{\textcolor{blue}{+0.03}}$ & $5.95^{\textcolor{blue}{+0.07}}$ & $9.28^{\textcolor{blue}{+0.16}}$ & $13.92^{\textcolor{blue}{+0.19}}$ & $4.38^{\textcolor{blue}{+0.45}}$  & $5.15^{\textcolor{blue}{+1.12}}$ \\
CNZSL(Tr) & 5.77 & 9.48 & 16.49 & 23.25 & 31.00 & 8.32 & 7.22 \\
$\quad$+HGR & $5.91^{\textcolor{blue}{+0.14}}$ & $9.54^{\textcolor{blue}{+0.06}}$ & $16.79^{\textcolor{blue}{+0.50}}$ & $23.83^{\textcolor{blue}{+0.58}}$ & $31.71^{\textcolor{blue}{+0.71}}$ & $8.56^{\textcolor{blue}{+0.24}}$ & $10.28^{\textcolor{blue}{+3.06}}$  \\
\hline
\end{tabular}
\end{adjustbox}
    \caption{First part explores our method with different loss, (e.g., - outer loss only calculates the inner loss). Second part compares CNZSL w/ or w/o additional graph knowledge. }
    \label{tab:compare}
\end{table}
\end{document}